%% file: main.tex
\documentclass[10pt,twocolumn,letterpaper]{article}

\usepackage{cvpr}              % To produce the CAMERA-READY version
\usepackage{graphicx}
\usepackage{amsmath}
\usepackage{amssymb}
\usepackage{booktabs}
\usepackage{times}
\usepackage{epsfig}
\usepackage{graphicx}
\usepackage{mmstyle}
\usepackage{pifont}% http://ctan.org/pkg/pifont
\usepackage{makecell}

\usepackage[pagebackref,breaklinks,colorlinks]{hyperref}

\usepackage[capitalize]{cleveref}
\crefname{section}{Sec.}{Secs.}
\Crefname{section}{Section}{Sections}
\Crefname{table}{Table}{Tables}
\crefname{table}{Tab.}{Tabs.}

\newcommand{\cmark}{\ding{51}}%
\newcommand{\xmark}{\ding{55}}%

\def\cvprPaperID{17} % *** Enter the CVPR Paper ID here
\def\confName{CVEU}
\def\confYear{2022}

\input{macros}

\begin{document}

\title{Temporal and Contextual Transformer for \\Multi-Camera Editing of TV Shows}

% 	Institution1\\
% 	Institution1 address\\
% 	{\tt\small firstauthor@i1.org}
% 	% For a paper whose authors are all at the same institution,
% 	% omit the following lines up until the closing ``}''.
% 	% Additional authors and addresses can be added with ``\and'',
% 	% just like the second author.
% 	% To save space, use either the email address or home page, not both
% 	\and
% 	Second Author\\
% 	Institution2\\
% 	First line of institution2 address\\
% 	{\tt\small secondauthor@i2.org}
% }

\author{Anyi Rao$^{2}$,
	Xuekun Jiang$^{3}$,
    Sichen Wang$^{1}$,
    Yuwei Guo$^{4}$,
    Zihao Liu$^{1}$,\\
    Bo Dai$^{3}$,
    Long Pang$^{1}$,
    Xiaoyu Wu$^{1}$,
    Dahua Lin$^{2,3}$,
    Libiao Jin$^{1}$\thanks{Corresponding author}\\
    $^{1}$Communication University of China \quad
    $^{2}$The Chinese University of Hong Kong\\
	$^{3}$Shanghai Artificial Intelligence Laboratory  \quad
	$^{4}$Nanjing University \\
	{\tt\small \{anyirao, dhlin\}@ie.cuhk.edu.hk}~
	{\tt\small \{jiangxuekun, daibo\}@pjlab.org.cn} \\
	{\tt\small guoyw@smail.nju.edu.cn}~
	{\tt\small \{kamino,panglong,wuxiaoyu,libiao\}@cuc.edu.cn}~
}

\maketitle

\input{articles/abstract.tex}

%%%%%%%%% BODY TEXT
\input{articles/introduction.tex}
\input{articles/related.tex}
\input{articles/dataset.tex}

\input{articles/methodology.tex}
\input{articles/experiment.tex}
\input{articles/conclusion.tex}

{\small
\bibliographystyle{ieee_fullname}
\bibliography{main}
}

\end{document}

%% file: macros.tex
\usepackage{xcolor}
\usepackage{wrapfig}
\usepackage{soul}

\definecolor{yellow}{rgb}{1,1, 0.6}
\definecolor{lightyellow}{rgb}{1,1, 0.8}
\definecolor{orange}{rgb}{1, 0.8, 0.6}
\definecolor{coral}{RGB}{246,131,65}
\definecolor{pinkred}{rgb}{1, 0.6, 0.6}
\definecolor{hotpink}{RGB}{238,64,195}
\definecolor{lavender}{RGB}{207,226,243}
\definecolor{gainsboro}{RGB}{208,224,227}
\definecolor{gainsboro2}{RGB}{217,234,211}
\definecolor{blanchedalmond}{RGB}{252,229,205}

%% file: articles/abstract.tex
% !TEX root = ../main.tex

\begin{abstract}
The ability to choose an appropriate camera view among multiple cameras plays a vital role in TV shows delivery. 
But it is hard to figure out the statistical pattern and apply intelligent processing due to the lack of high-quality training data. 
To solve this issue, we first collect a novel benchmark on this setting with four diverse scenarios including concerts, sports games, gala shows, and contests, where each scenario contains 6 synchronized tracks recorded by different cameras. 
It contains 88-hour raw videos that contribute to the 14-hour edited videos. Based on this benchmark, we further propose a new approach temporal and contextual transformer that utilizes clues from historical shots and other views to make shot transition decisions and predict which view to be used. Extensive experiments show that our method outperforms existing methods on the proposed multi-camera editing benchmark.~\footnote{A shot is a series of continuous frames recorded by a camera and a track refers to the video recorded by one camera from a specific view. }
\end{abstract}

%% file: articles/introduction.tex
% !TEX root = ../main.tex

\section{Introduction}
\label{sec:introduction}

Multiview cameras provide more angles and framing in videos which help the viewer to relate with the story and performance in a much better way. Though they are more visually compelling than single-view videos, their video editing process is more tedious since it requires an experienced editor to switch among different tracks to guide the viewer.

Although intelligent video editing and creation span across shot type classification~\cite{rao2020unified}, cut point ranking~\cite{pardo2021learning}, cut type recognition~\cite{pardo2022moviecuts}, transition type recommendation~\cite{shen2022autotransition}, \etc,
there are few works working on multiview video computational editing, 
which aims to present a normal view video with shots changes automatically.
The first step towards this goal is to select an appropriate view from the tracks of multiview cameras at each time step.
The challenges come from two aspects:
to select a suitable cutting point that transits from one shot to another shot and ensures the temporal consistency among shots in the final edited video; 
to pick up an appropriate view from candidate views that can well present the whole plot.

\begin{figure}[!t]
	\centering
	\includegraphics[width=\linewidth]{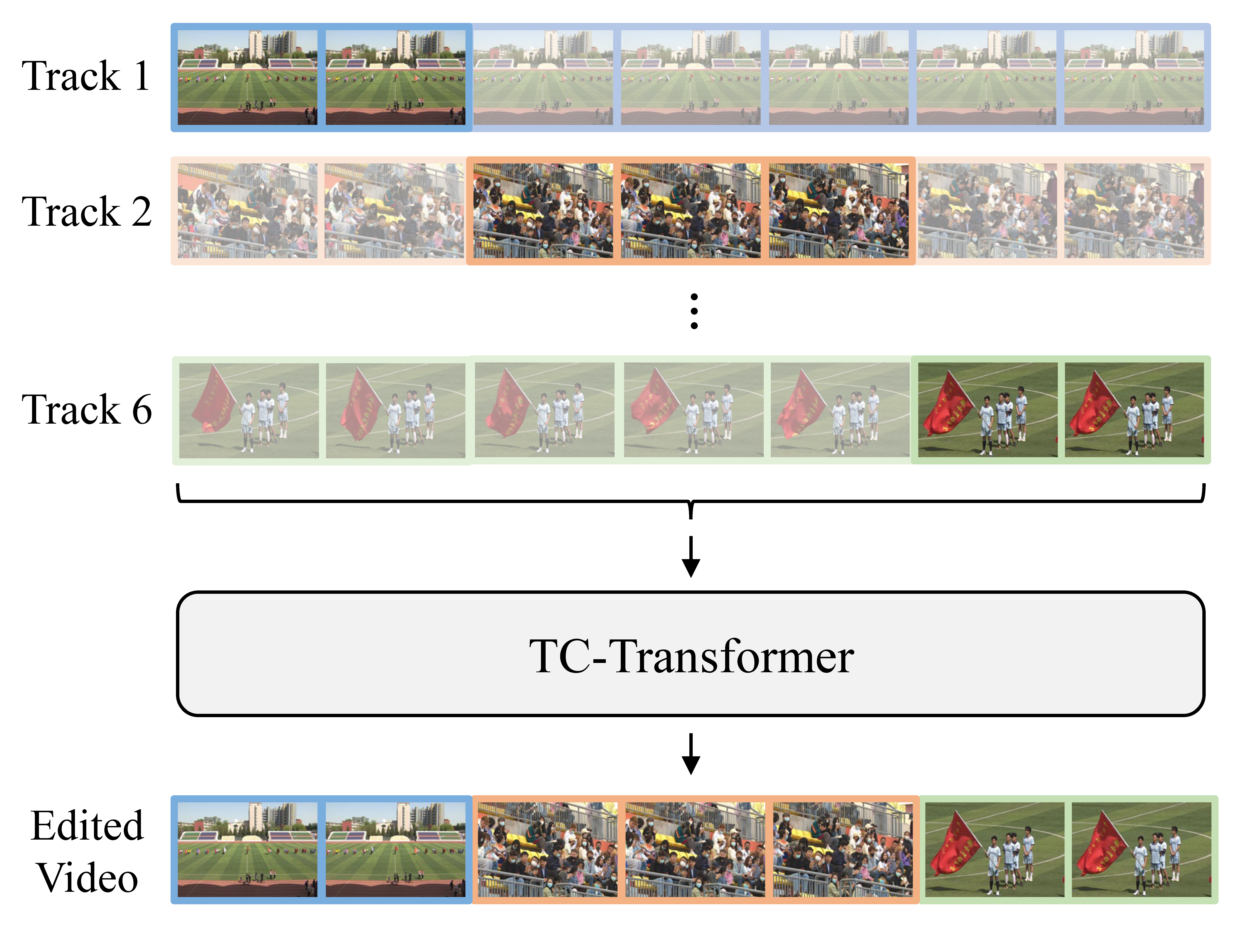}
	\vspace{-20pt}
	\caption{Multi-camera editing from 6 tracks synchronized raw footage using our proposed TC-Transformer.}
 	\vspace{-8pt}
	\label{fig:teaser}
\end{figure}

% Though there is no existing method that handles the exact same setting as ours, to the best of our knowledge, relevant approaches on multiview video summarization~\cite{} can be adapted to this setting. 
Existing methods~\cite{hussain2019cloud, meng2017video}  basically rely on manually crafted rules or are trained on small datasets with a limited number of samples to tackle the surveillance camera scenarios.
The lack of large-scale diverse training data restrains it from real-world applications in multi-camera video editing.
To address this limitation,
we focus on a common application domain TV shows and collect a large-scale dataset with 6 synchronized tracks of raw videos and ground truth videos from professional TV shows.
It covers indoor and outdoor scenarios spanning concerts, sports games, gala shows and contests and lasts for 88 hours which is 50 times larger than existing datasets.
Taking the advantage of the above dataset, 
our proposed Temporal and Contextual Transformer (TC-Transformer) integrates the cutting point selection and track prediction into a unified framework that utilizes a sliding window to classify the correct view.
It helps to achieve $5\%\sim6\%$ relative improvement over baselines.

To this end, our contribution can be summarized as follows,
1) a novel multi-camera editing dataset that contains 6 raw synchronized tracks in 88 hours recorded by professionals for the ground truth edited video made by experienced video editors.
2) a target designed approach temporal and contextual transformer that utilizes the clues of historical shots and concurrent videos from other views
and solves the cutting point and view selection jointly.

\vspace{2pt}
\noindent\textbf{Assumption}: Although the paper focuses on the shot transition and view selection for TV shows,
the transition effects, such as fade in/out, among different shots, which can bring more advanced performance to the final edited video, its extension to other application domains, \eg, daily life videos,
are out of the scope of this paper.

%% file: articles/related.tex
% !TEX root = ../main.tex

\section{Related Work}
\label{sec:related}
\noindent\textbf{Shot-level Video Editing.}
Besides pixel-level editing~\cite{huynh2021new,rao2022coarse,li2022towards,interactiveDepth},
intelligent shot-level video editing tools also help users to create their videos more efficiently.
Some approaches tackle the textual semantics of the generated video~\cite{fried2019text,huber2019b,truong2016quickcut,pavel2015sceneskim,berthouzoz2012tools,huang2020movienet}.
Wang~\etal~\cite{wang2019write} automatically remix semantic-matched shots based on the text entered by the user.
Xiong~\etal~\cite{xiong2021transcript} develop a weakly-supervised framework that uses text as input to automatically create video sequences from a shot collection. 
Some researchers focus on generating various video styles based on various manually defined conditions~\cite{lino2015intuitive,lino2011computational,merabti2016virtual,arev2014automatic,gandhi2014multi,rao2022shoot360}.
Leaken~\etal~\cite{leake2017computational} propose a system for efficient video editing by offering a set of basic idioms.
Frey~\etal~\cite{frey2021automatic} develop an automatic approach that extracts editing styles in a source video and applies them to corresponding footage.
Other works study the attributes and connection of individual shots~\cite{argaw2022anatomy,pardo2022moviecuts,pardo2021learning}.
It appears a rich set of benchmarks on shot sequence ordering, next shot selection, missing shot attributes prediction~\cite{argaw2022anatomy}, cut prediction~\cite{pardo2022moviecuts}, \etc.
In this paper, we focus on how to create videos from multi-camera footage with the aim of following the consistency of temporal and contextual relationships.

\begin{table}[!t]
	\begin{center}
		\caption{Comparison with existing multiview video summarization dataset.}
		\vspace{-4pt}
 		\resizebox{\linewidth}{!}{%
			\begin{tabular}{c|c|c|c}
                \toprule
				   &  Sync.  & Scenarios &  Duration(h)   \\ \midrule
            Surveillance~\cite{fu2010multi}    &  \xmark  & \makecell{office, campus,\\ road, badminton}  & 1.75    \\ \midrule 
            Social~\cite{arev2014automatic}    &  \xmark  & \makecell{office, park, \\ street, snowman}  & 1.01    \\ \midrule 
            Multi-Camera Editing &  \cmark  &  \makecell{concert, context\\ sports, gala show}   &  \textbf{88.44}   \\
 				 \bottomrule
			\end{tabular}
 		}
 		\vspace{-10pt}
		\label{tab:dataset_comp}
	\end{center}
\end{table}

\vspace{4pt}
\noindent\textbf{Video Summarization.}
Another branch of work that shares similarities with ours is video summarization~\cite{apostolidis2021video,hussain2021comprehensive} and video highlight detection~\cite{wang2020learning} in the format of "select video clips from a large set of footage",
with the main purpose of compressing information.
Among the above, multiview video summarization~\cite{hussain2021comprehensive} is the most related to ours.
Most methods~\cite{hussain2021comprehensive} study the surveillance scenario~\cite{fu2010multi},
and a few of them focus on social events~\cite{arev2014automatic}.
A brief comparison of these datasets and our multi-camera editing dataset is shown in Table~\ref{tab:dataset_comp}.
The advantages of our new dataset lie in its synchronization, long length and clear targets to TV shows multi-camera editing.
With the dataset, Hussain~\etal\cite{hussain2019cloud} proposed a DB-LSTM based approach to generate a summary from multiview videos,
which utilizes a lookup table to save computational resources.
Meng~\etal~\cite{meng2017video} formulate video summarization as a multiview representative selection problem that aims to find a selection of visual elements agreeable with all views.
However,
all of the existing approaches 
target saliency detection by choosing the most noticeable parts from the video footage.
Instead, 
our multi-camera editing aims to pick up one ideal clips from different tracks at the same timestamp in much more challenging scenarios such as concerts, gala shows, sports activities.
The editing is required to achieve consistency in consecutive shots to deliver a coherent long video~\cite{jacobson2012mastering}.

%% file: articles/dataset.tex
% !TEX root = ../main.tex

\section{TV Shows Multi-Camera Editing Dataset}
\label{sec:dataset}
\begin{figure}[!t]
	\centering
	\includegraphics[width=\linewidth]{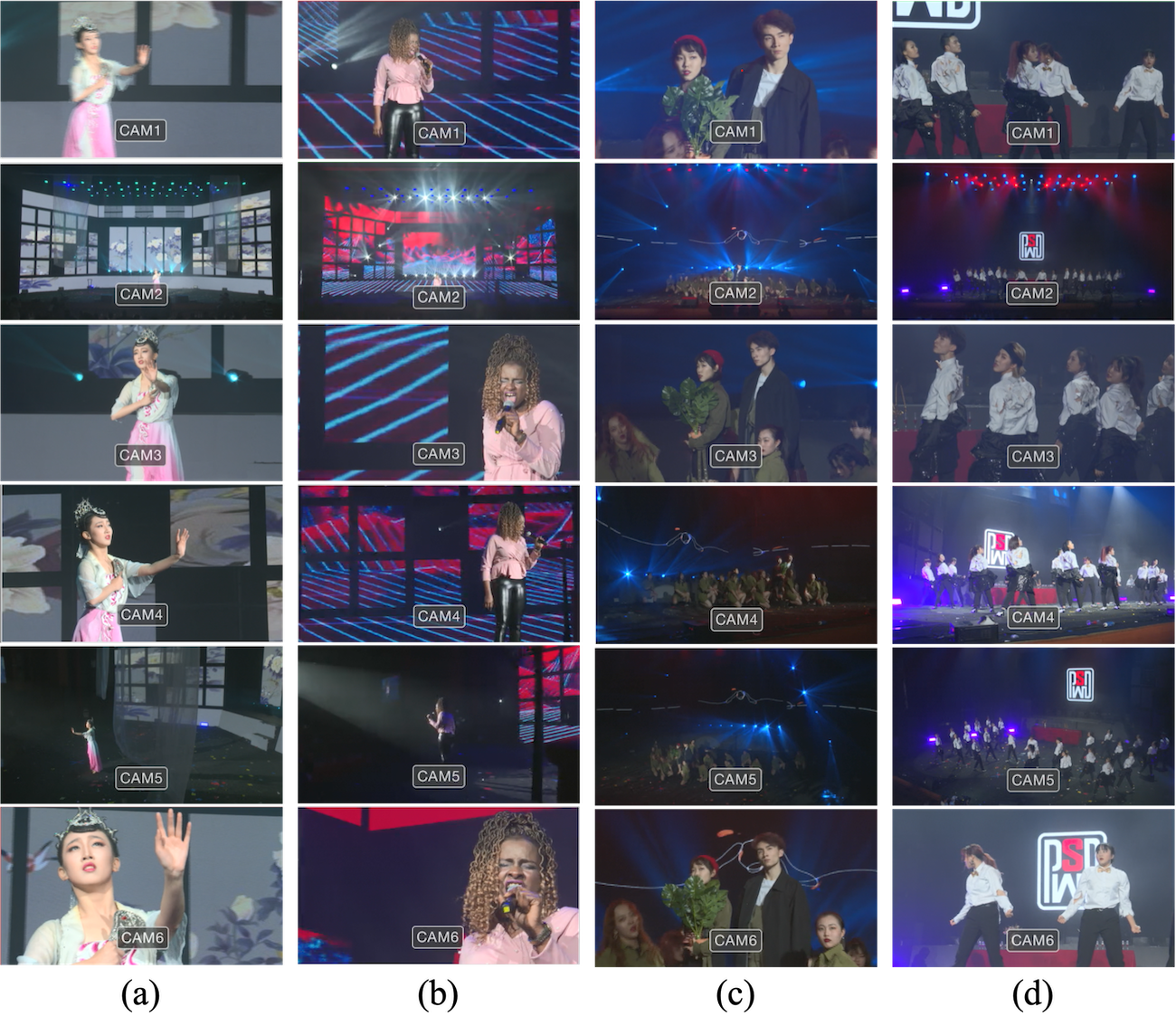}
	\vspace{-20pt}
	\caption{A glimpse of the TVMCE dataset on four different scenarios with six different synchronized tracks.}
	\vspace{5pt}
	\label{fig:data}
\end{figure}

To facilitate the research in multi-camera editing,
we collect a large-scale high-quality dataset TV shows Multi-Camera Editing (TVMCE) dataset,
with a glimpse of it shown in Fig.~\ref{fig:data}.
%A demo video of the dataset is shown in the supplementary materials and we encourage readers to watch it.

\begin{figure}[!t]
	\centering
	\includegraphics[width=\linewidth]{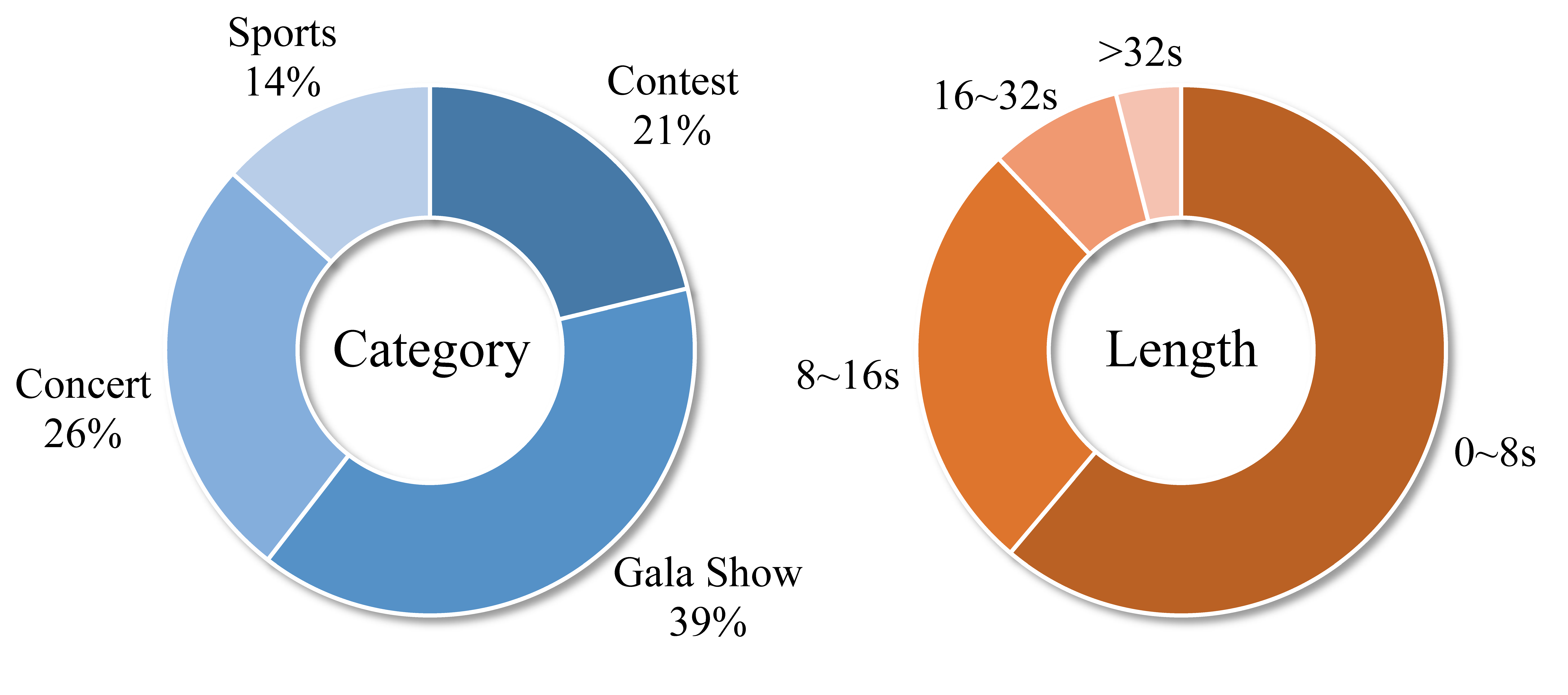}
	\vspace{-20pt}
	\caption{\emph{Left:} Proportion of each shot category in the TVMCE dataset; 
	\emph{Right:} the shot length distribution of edited videos.}
	\vspace{5pt}
	\label{fig:stat}
\end{figure}

\vspace{4pt}
\noindent\textbf{Data Collection and Procedure.}
To make a high-quality multi-camera editing dataset for TV shows,
its content, videography and editing should reach a professional level. 
With this in mind, we reached out to 
film and TV production major colleges and follow their professional production team to acquire data covering various scenarios including concert, sports, contest and gala show that happen in universities and city theaters/stadiums.
These provide real-world content for us to study.
With the physical content, to professionally record them is also of great importance.
The videographers are professionally trained senior final-year student who are guided by professors as directors.
After collecting the raw footage,
The professional director need to check all raw footage and decide when to use the footage and for how long based on the specific content of the performance and some professional knowledge.

The collection and annotation of the TVMCE dataset require a huge cost of manpower and time with two key challenges.
The first one lies on synchronization, that is to align the timestamps of each view track and every camera takes the same content from different views at any given moment. 
The second challenge is how to efficiently select the right track from the multiple view tracks. 
For an event, \eg, a concert or a gala show, its usually takes 3$\sim$4 hours. 
This means that the director needs more than 20 hours to watch these videos, which is unacceptable. 
To ease the above difficulties,
all different view tracks are pre-synchronized in camera settings and fed into the same screen. 
The editor can see all the tracks at once and make their decision, and the results are double checked by other professionals to ensure the quality. 
By recording the actions of these directors, we can get the annotation of which and when a track is selected.

\vspace{4pt}
\noindent\textbf{Statistics.}
As shown in Figure~\ref{fig:stat},
our dataset holds a balanced coverage ratio among different scenario categories with 39\% in gala shows and 14\% in sports.
Most shots in our dataset have a time duration between 0 to 8 seconds and a few shots are long shots that last longer than 32 seconds.

%% file: articles/methodology.tex
% !TEX root = ../main.tex
\section{Temporal and Contextual Transformer}
\label{sec:methodology}

\noindent\textbf{Problem Formulation.}
The raw input for multi-camera editing is the video pool $V$ that lasts for $I$ frames accompanying with $J$ tracks.
$$
V = \begin{bmatrix}
v_{1,1} & v_{2,1} & v_{3,1}&\cdots & v_{I,1}\\
v_{1,2} & v_{2,2} & v_{2,2} &\cdots & v_{I,2}\\
\vdots & \vdots & \vdots & v_{i,j} & \vdots \\
v_{1,J} & v_{2,J} & v_{3,J} &\cdots & v_{I,J}\\
\end{bmatrix},
$$
where $v_{i,j}$ denotes the $j$-th track of $i$-th frame.
Our goal is to find a optimal combination of $v_{i,j}$ to form a video $\hat{V}$ such that it is consistent and rich in the content delivery.

\vspace{4pt}
\noindent\textbf{Methodology.}
The brute force way is to enumerate all the possibility in the complexity of $O(I^J)$ and it has to be done offline.
To overcome the above issues, we formulate this problem as a classification problem in a sliding window of length as $w$ to make a prediction $p_{i,j}$ for $v_{i,j}$ to determine whether it is appropriate to be presented in the final output.

\begin{figure}[t]
	\centering
	\includegraphics[width=0.85\linewidth]{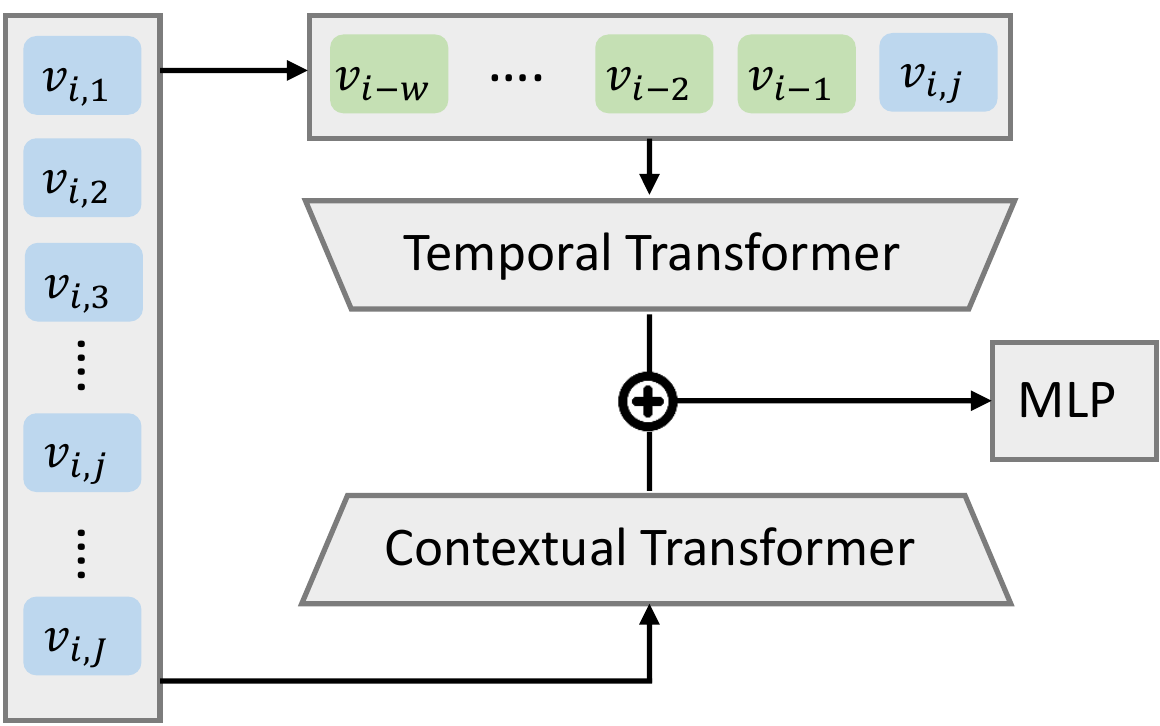}
	\vspace{-1pt}
	\caption{The framework of TC-Transformer. It contains a temporal transformer to take in historical video frames including $i-w$-th frame to $i$-th frame and a contextual transformer to process all the possible candidate frames at $i$-th timestamp.}
	\vspace{3pt}
	\label{fig:pipeline}
\end{figure}

Inspired by the observation that professional editors select a view based on the historical decision they made and all the available views at the current time stamp.
To fully utilize the above properties, 
we design two transformers on each frame $v_{i,j}$ to deal with temporal consistency and contextual coherence.
Specifically, the temporal transformer $\cT$ takes $v_{i,j}$ and $i-w$-th to $i-1$-th frames as input.
The contextual transformer $\cC$ incorporates the information from all the frames $v_{i,1}$ to $v_{i,J}$ at $i$-th timestamp.
The outcomes of the above two transformers are finally fed into a MLP layer and the process can be formulated as below,
\begin{equation}
\begin{split}
p_{i,j} = \cF(&\cT(v_{i-w}, \cdots, v_{i-1}, v_{i,j}) \\
&\oplus \cC([v_{i,1}, \cdots, v_{i,j}, \cdots, v_{i,J})),
\end{split}
\end{equation}
where $p_{i,j} \in [0,1]$ is the predication score of whether $v_{i,j}$ will be selected.
The whole framework is trained end-to-end with binary classification loss,
\begin{equation} 
\cL = - y \log (p) + (1-y) \log(1-p),
\end{equation}
where $y$ is the ground truth, \ie, $y=1$ indicates $v_{i,j}$ is the selected as $v_i$, vice versa.

\vspace{4pt}
\noindent\textbf{Efficient Training.}
The aforementioned training requires generating samples at each time step for every frame that requires heavy computational resources, 
since there are 5.8M frames for 88-hour footage with the fps as 24.~\footnote{5.8M =$66.9\times3,600\times24$, that is all the training frames.}
And it also casts difficulties in learning useful information related to multi-camera editing.
To speed up the training process, we focus on shot boundaries and generate training samples using a window size as 16 and a step size as 5 frames.
This reduces the training samples to 24K.~\footnote{24K=$4,042\times6$, that is 6 samples for each shot boundary.}

%% file: articles/experiment.tex
% !TEX root = ../main.tex

\section{Experiments}
\label{sec:experiment}

\subsection{Setup}
\noindent\textbf{Evaluation.}
All the baseline methods are tested on our
TV Shows Multi-Camera Editing (TVMCE) dataset,
which is split into 4:1 for training and test as shown in Figure~\ref{tab:dataset}.
To evaluate the performance, we calculate the Precision and the Average Precision (AP) of selecting the right track, 
where the Precision is calculated by taking a concrete threshold $0.5$ and the Average Precision is calculated as the weighted mean of precisions at each threshold.

\vspace{4pt}
\noindent\textbf{Baseline.}
TSN~\cite{wang2016temporal} models long-range temporal structures with a segment-based sampling and aggregation module.
SlowFast~\cite{feichtenhofer2019slowfast}
is composed of a Slow pathway operating at low frame rate that captures spatial semantics, and a Fast pathway operating at high frame rate that  captures motion at fine temporal resolution. 
ViVit~\cite{arnab2021vivit}
extracts spatio-temporal tokens from the input video and encodes them into a series of transformer layers.
Video-Swin~\cite{liu2022video}
advocates an inductive bias of locality and computes self-attention globally with spatial-temporal factorization.

\vspace{4pt}
\noindent\textbf{Implementation Details.}
Both TSN~\cite{wang2016temporal} and SlowFast~\cite{feichtenhofer2019slowfast} use ResNet50~\cite{he2016deep} as the backbone, 
which is pretrained on the ImageNet dataset~\cite{deng2009imagenet}. 
The ViVit~\cite{arnab2021vivit}
is consisted of a spatial transformer and a temporal transformer, which use 12 transformer encoder layers. 
Each layer has a self-attention block of 12 heads with hidden dimension as 768. 
For Video-Swin~\cite{liu2022video}, 
the dimension of the hidden layers in the first stage is set to 96, 
and 
the layer numbers of transformer block in each stage is 2, 2, 6, 2. 
The patch size is set to 2x4x4 and the window size is 8x7x7. Both ViVit and Video-Swin are pretrained with Kinetics-400~\cite{kay2017kinetics}. 
For a fair comparison, all methods use a MLP block to calculate the classification score after layer normalization with the hidden dimension as 192.

\begin{table}[!t]
	\begin{center}
		\caption{Data split of the TVMCE Dataset.}
		\vspace{-4pt}
% 		\resizebox{0.7\linewidth}{!}{%
			\begin{tabular}{c|c|c|c}
                \toprule
				   &  Train  &  Test   &  Total   \\ \midrule
            \# of GT shots      &  4,042  &  1,091   &  5,133   \\ 
            Duration(h) &  66.90  &  21.54   &  88.44   \\
            % Duration(h) &  11.15  &  3.59   &  14.74   \\
 				 \bottomrule
			\end{tabular}
% 		}
		\vspace{-5pt}
		\label{tab:dataset}
	\end{center}
\end{table}

\subsection{Overall Analysis}
The overall results are shown in Table~\ref{tab:results}.
It is observed that with the training on TVMCE dataset,
TSN acquires $16.88$ on precision and $18.70$ on AP,
which performs better than a random guess.
As the backbone becomes deeper and more powerful in integrating temporal information, ViVit~\cite{arnab2021vivit} and Cloud~\cite{hussain2019cloud}
get
$1\%\sim3\%$ improvement on precision and AP.
Due to the advantage of hierarchy and locality representation ability,
Video-Swin~\cite{liu2022video} improves the precision from $20.38$ to $21.20$ and the AP from $20.38$ to $20.44$, which
outperforms the above.

Benefited from the joint usage of temporal and contextual information, 
TC-Transformer achieves the highest precision accuracy $22.48$ and AP $21.62$ on the test set of TVMCE dataset, 
with $5\%\sim6\%$ relative improvement over the baseline.

\begin{table}[!t]
	\begin{center}
		\caption{Overall results on TVMCE Dataset.}
		\vspace{-4pt}
% 		\resizebox{0.92\linewidth}{!}{%
			\begin{tabular}{c|c|c}
                \toprule
				Method    &  Precision(\%)    &  AP(\%)   \\ \midrule
                Random      &    16.66       &    16.66 \\ 
                TSN~\cite{wang2016temporal}       &    16.88        &    18.70 \\
                SlowFast~\cite{feichtenhofer2019slowfast}       &  16.70           &    16.67 \\  
                ViVit~\cite{arnab2021vivit}       &    17.60      &    19.83 \\
                Cloud~\cite{hussain2019cloud}       &    20.38   &    20.38 \\ 
                Video-Swin~\cite{liu2022video}       &    21.20        &    20.44 \\  \midrule
                TC-Transformer(ViVit)       &    21.29         &    20.15 \\
                TC-Transformer(Swin)       &    \textbf{22.48}        &    \textbf{21.62} \\
 				 \bottomrule
			\end{tabular}
%  		}
 		\vspace{-15pt}
		\label{tab:results}
	\end{center}
\end{table}

%% file: articles/conclusion.tex
% !TEX root = ../main.tex

\section{Conclusion}
\label{sec:conclusion}
In this paper, we propose a new TV shows Multi-Camera Editing (TVMCE) dataset for multi-camera editing. 
It contains 88-hour raw synchronized footage recorded by 6 different cameras from different views and covers four main type of TV shows including sports, gala shows, concerts, contests.
We further introduce a new method temporal and contextual transformer TC-Transformer that 
selects the desired frame
according to its historical frames and concurrent frames from different views.
The comparison on TVMCE dataset with state-of-the-art baselines shows the superiority of our TC-Transformer.

% \noindent\textbf{Acknowledgement}: 
% This work is supported by 2021YFF0900701

%% file: main.bbl
\begin{thebibliography}{10}\itemsep=-1pt

\bibitem{apostolidis2021video}
Evlampios Apostolidis, Eleni Adamantidou, Alexandros~I Metsai, Vasileios
  Mezaris, and Ioannis Patras.
\newblock Video summarization using deep neural networks: A survey.
\newblock {\em Proceedings of the IEEE}, 109(11):1838--1863, 2021.

\bibitem{arev2014automatic}
Ido Arev, Hyun~Soo Park, Yaser Sheikh, Jessica Hodgins, and Ariel Shamir.
\newblock Automatic editing of footage from multiple social cameras.
\newblock {\em ACM Transactions on Graphics (TOG)}, 33(4):1--11, 2014.

\bibitem{argaw2022anatomy}
Dawit~Mureja Argaw, Fabian~Caba Heilbron, Joon-Young Lee, Markus Woodson, and
  In~So Kweon.
\newblock The anatomy of video editing: A dataset and benchmark suite for
  ai-assisted video editing.
\newblock In {\em The European Conference on Computer Vision (ECCV)}, 2022.

\bibitem{arnab2021vivit}
Anurag Arnab, Mostafa Dehghani, Georg Heigold, Chen Sun, Mario Lu{\v{c}}i{\'c},
  and Cordelia Schmid.
\newblock Vivit: A video vision transformer.
\newblock {\em arXiv preprint arXiv:2103.15691}, 2021.

\bibitem{berthouzoz2012tools}
Floraine Berthouzoz, Wilmot Li, and Maneesh Agrawala.
\newblock Tools for placing cuts and transitions in interview video.
\newblock {\em ACM Transactions on Graphics (TOG)}, 31(4):1--8, 2012.

\bibitem{deng2009imagenet}
Jia Deng, Wei Dong, Richard Socher, Li-Jia Li, Kai Li, and Li Fei-Fei.
\newblock Imagenet: A large-scale hierarchical image database.
\newblock In {\em 2009 IEEE conference on computer vision and pattern
  recognition}, pages 248--255. Ieee, 2009.

\bibitem{interactiveDepth}
Obumneme~Stanley Dukor, S.~Mahdi~H. Miangoleh, Mahesh Kumar~Krishna Reddy, Long
  Mai, and Ya\u{g}{\i}z Aksoy.
\newblock Interactive editing of monocular depth.
\newblock In {\em SIGGRAPH Posters}, 2022.

\bibitem{feichtenhofer2019slowfast}
Christoph Feichtenhofer, Haoqi Fan, Jitendra Malik, and Kaiming He.
\newblock Slowfast networks for video recognition.
\newblock In {\em Proceedings of the IEEE/CVF international conference on
  computer vision}, pages 6202--6211, 2019.

\bibitem{frey2021automatic}
Nathan Frey, Peggy Chi, Weilong Yang, and Irfan Essa.
\newblock Automatic non-linear video editing transfer.
\newblock {\em arXiv preprint arXiv:2105.06988}, 2021.

\bibitem{fried2019text}
Ohad Fried, Ayush Tewari, Michael Zollh{\"o}fer, Adam Finkelstein, Eli
  Shechtman, Dan~B Goldman, Kyle Genova, Zeyu Jin, Christian Theobalt, and
  Maneesh Agrawala.
\newblock Text-based editing of talking-head video.
\newblock {\em ACM Transactions on Graphics (TOG)}, 38(4):1--14, 2019.

\bibitem{fu2010multi}
Yanwei Fu, Yanwen Guo, Yanshu Zhu, Feng Liu, Chuanming Song, and Zhi-Hua Zhou.
\newblock Multi-view video summarization.
\newblock {\em IEEE Transactions on Multimedia}, 12(7):717--729, 2010.

\bibitem{gandhi2014multi}
Vineet Gandhi, Remi Ronfard, and Michael Gleicher.
\newblock Multi-clip video editing from a single viewpoint.
\newblock In {\em Proceedings of the 11th European Conference on Visual Media
  Production}, pages 1--10, 2014.

\bibitem{he2016deep}
Kaiming He, Xiangyu Zhang, Shaoqing Ren, and Jian Sun.
\newblock Deep residual learning for image recognition.
\newblock In {\em Proceedings of the IEEE conference on computer vision and
  pattern recognition}, pages 770--778, 2016.

\bibitem{huang2020movienet}
Qingqiu Huang, Yu Xiong, Anyi Rao, Jiaze Wang, and Dahua Lin.
\newblock Movienet: A holistic dataset for movie understanding.
\newblock In {\em European Conference on Computer Vision}, pages 709--727.
  Springer, 2020.

\bibitem{huber2019b}
Bernd Huber, Hijung~Valentina Shin, Bryan Russell, Oliver Wang, and Gautham~J
  Mysore.
\newblock B-script: Transcript-based b-roll video editing with recommendations.
\newblock In {\em Proceedings of the 2019 CHI Conference on Human Factors in
  Computing Systems}, pages 1--11, 2019.

\bibitem{hussain2021comprehensive}
Tanveer Hussain, Khan Muhammad, Weiping Ding, Jaime Lloret, Sung~Wook Baik, and
  Victor Hugo~C de Albuquerque.
\newblock A comprehensive survey of multi-view video summarization.
\newblock {\em Pattern Recognition}, 109:107567, 2021.

\bibitem{hussain2019cloud}
Tanveer Hussain, Khan Muhammad, Amin Ullah, Zehong Cao, Sung~Wook Baik, and
  Victor Hugo~C de Albuquerque.
\newblock Cloud-assisted multiview video summarization using cnn and
  bidirectional lstm.
\newblock {\em IEEE Transactions on Industrial Informatics}, 16(1):77--86,
  2020.

\bibitem{huynh2021new}
Loc Huynh, Bipin Kishore, and Paul Debevec.
\newblock A new dimension in testimony: Relighting video with reflectance field
  exemplars.
\newblock {\em arXiv preprint arXiv:2104.02773}, 2021.

\bibitem{jacobson2012mastering}
Mitch Jacobson.
\newblock {\em Mastering Multi-Camera Techniques: From Pre-Production to
  Editing to Deliverable Masters}.
\newblock CRC Press, 2012.

\bibitem{kay2017kinetics}
Will Kay, Joao Carreira, Karen Simonyan, Brian Zhang, Chloe Hillier, Sudheendra
  Vijayanarasimhan, Fabio Viola, Tim Green, Trevor Back, Paul Natsev, et~al.
\newblock The kinetics human action video dataset.
\newblock {\em arXiv preprint arXiv:1705.06950}, 2017.

\bibitem{leake2017computational}
Mackenzie Leake, Abe Davis, Anh Truong, and Maneesh Agrawala.
\newblock Computational video editing for dialogue-driven scenes.
\newblock {\em ACM Trans. Graph.}, 36(4):130--1, 2017.

\bibitem{li2022towards}
Zhen Li, Cheng-Ze Lu, Jianhua Qin, Chun-Le Guo, and Ming-Ming Cheng.
\newblock Towards an end-to-end framework for flow-guided video inpainting.
\newblock In {\em Proceedings of the IEEE/CVF Conference on Computer Vision and
  Pattern Recognition}, pages 17562--17571, 2022.

\bibitem{lino2011computational}
Christophe Lino, Mathieu Chollet, Marc Christie, and R{\'e}mi Ronfard.
\newblock Computational model of film editing for interactive storytelling.
\newblock In {\em International Conference on Interactive Digital
  Storytelling}, pages 305--308. Springer, 2011.

\bibitem{lino2015intuitive}
Christophe Lino and Marc Christie.
\newblock Intuitive and efficient camera control with the toric space.
\newblock {\em ACM Transactions on Graphics (TOG)}, 34:1--12, 2015.

\bibitem{liu2022video}
Ze Liu, Jia Ning, Yue Cao, Yixuan Wei, Zheng Zhang, Stephen Lin, and Han Hu.
\newblock Video swin transformer.
\newblock In {\em Proceedings of the IEEE/CVF Conference on Computer Vision and
  Pattern Recognition}, pages 3202--3211, 2022.

\bibitem{meng2017video}
Jingjing Meng, Suchen Wang, Hongxing Wang, Junsong Yuan, and Yap-Peng Tan.
\newblock Video summarization via multi-view representative selection.
\newblock In {\em Proceedings of the IEEE International Conference on Computer
  Vision Workshops}, pages 1189--1198, 2018.

\bibitem{merabti2016virtual}
Billal Merabti, Marc Christie, and Kadi Bouatouch.
\newblock A virtual director using hidden markov models.
\newblock In {\em Computer Graphics Forum}, volume~35, pages 51--67. Wiley
  Online Library, 2016.

\bibitem{pardo2021learning}
Alejandro Pardo, Fabian Caba, Juan~Le{\'o}n Alc{\'a}zar, Ali~K Thabet, and
  Bernard Ghanem.
\newblock Learning to cut by watching movies.
\newblock In {\em Proceedings of the IEEE/CVF International Conference on
  Computer Vision}, pages 6858--6868, 2021.

\bibitem{pardo2022moviecuts}
Alejandro Pardo, Fabian~Caba Heilbron, Juan~Le{\'o}n Alc{\'a}zar, Ali Thabet,
  and Bernard Ghanem.
\newblock Moviecuts: A new dataset and benchmark for cut type recognition.
\newblock In {\em The European Conference on Computer Vision (ECCV)}, 2022.

\bibitem{pavel2015sceneskim}
Amy Pavel, Dan~B Goldman, Bj{\"o}rn Hartmann, and Maneesh Agrawala.
\newblock Sceneskim: Searching and browsing movies using synchronized captions,
  scripts and plot summaries.
\newblock In {\em Proceedings of the 28th Annual ACM Symposium on User
  Interface Software \& Technology}, pages 181--190, 2015.

\bibitem{rao2020unified}
Anyi Rao, Jiaze Wang, Linning Xu, Xuekun Jiang, Qingqiu Huang, Bolei Zhou, and
  Dahua Lin.
\newblock A unified framework for shot type classification based on subject
  centric lens.
\newblock In {\em The European Conference on Computer Vision (ECCV)}, 2020.

\bibitem{rao2022coarse}
Anyi Rao, Linning Xu, Zhizhong Li, Qingqiu Huang, Zhanghui Kuang, Wayne Zhang,
  and Dahua Lin.
\newblock A coarse-to-fine framework for automatic video unscreen.
\newblock {\em IEEE Transactions on Multimedia}, 2022.

\bibitem{rao2022shoot360}
Anyi Rao, Linning Xu, and Dahua Lin.
\newblock Shoot360: Normal view video creation from city panorama footage.
\newblock In {\em ACM SIGGRAPH 2022 Conference Proceedings}, pages 1--9, 2022.

\bibitem{shen2022autotransition}
Yaojie Shen, Libo Zhang, Kai Xu, and Xiaojie Jin.
\newblock Autotransition: Learning to recommend video transition effects.
\newblock In {\em The European Conference on Computer Vision (ECCV)}, 2022.

\bibitem{truong2016quickcut}
Anh Truong, Floraine Berthouzoz, Wilmot Li, and Maneesh Agrawala.
\newblock Quickcut: An interactive tool for editing narrated video.
\newblock In {\em Proceedings of the 29th Annual Symposium on User Interface
  Software and Technology}, pages 497--507, 2016.

\bibitem{wang2020learning}
Lezi Wang, Dong Liu, Rohit Puri, and Dimitris~N Metaxas.
\newblock Learning trailer moments in full-length movies with co-contrastive
  attention.
\newblock In {\em European Conference on Computer Vision}, pages 300--316.
  Springer, 2020.

\bibitem{wang2016temporal}
Limin Wang, Yuanjun Xiong, Zhe Wang, Yu Qiao, Dahua Lin, Xiaoou Tang, and
  Luc~Van Gool.
\newblock Temporal segment networks: Towards good practices for deep action
  recognition.
\newblock In {\em European conference on computer vision}, pages 20--36.
  Springer, 2016.

\bibitem{wang2019write}
Miao Wang, Guo-Wei Yang, Shi-Min Hu, Shing-Tung Yau, and Ariel Shamir.
\newblock Write-a-video: computational video montage from themed text.
\newblock {\em ACM Trans. Graph.}, 38(6):177--1, 2019.

\bibitem{xiong2021transcript}
Yu Xiong, Fabian~Caba Heilbron, and Dahua Lin.
\newblock Transcript to video: Efficient clip sequencing from texts.
\newblock {\em arXiv preprint arXiv:2107.11851}, 2021.

\end{thebibliography}
